\pdfoutput=1
\documentclass[11pt,twocolumn,a4paper]{article}
\usepackage{times}
\usepackage{latexsym}
\usepackage{xcolor}  
\usepackage{graphicx}  
\usepackage{url}
\usepackage{authblk}
\usepackage{microtype}

\usepackage{tikz}
\usepackage{bm}

\title{Contextualized Word Vector-based Methods for Discovering Semantic Differences with No Training nor Word Alignment}

\author[1]{Ryo Nagata}
\author[2]{Hiroya Takamura}
\author[3]{Noki Otani}
\author[4]{Yoshifumi Kawasaki}
\affil[1]{Konan University}
\affil[2]{ National Institute of Advanced Industrial Science and Technology}
\affil[3]{Tokyo University of Foreign Studies}
\affil[4]{University of Tokyo}

\date{}

\begin{document}
\maketitle
\section{Introduction}\label{sec:intro}
In this paper, we show that norms of contextualized word vectors obtained from a large language model are a good indicator for words exhibiting semantic differences\footnote{Following the convention in the literature, we use the term \textit{semantic difference} rather abstractly to refer to differences in meaning and usages.} in two corpora. To be precise, we show that the more meanings a word covers in a corpus, the shorter the norm of its mean word vector gets. Using this property, we propose methods for detecting semantic differences with their instances in context, which brings out various applications: e.g., second language acquisition research~\cite{mcenery} (e.g., words and their meanings non-native speakers do not use as native speakers), social linguistics~\cite{lei} (e.g., revealing semantic differences of words between British and American English), and historical linguistics~\cite{hamilton-etal-2016-diachronic} (e.g., discovering words that have acquired a meaning).

The major approach to semantic difference detection, which is based on non-contextualized word vectors such as Word2vec~\cite{mikolov2013distributed},
has several limitations and thus is not always applicable to any corpora as will be discussed in detail in Sect.\,\ref{sec:related_work}. Above all, many of non-contextualized word vector-based methods require some sort of correspondence between two corpora for comparison (e.g., word alignment). The task is, however, to find words that do not correspond well in terms of their meanings, and thus it is more natural not to assume any correspondence in advance as \cite{aida-etal-2021-comprehensive} point out. For example, it is not straightforward at all to align words between native and non-native English corpora. Besides, most previous methods are computationally costly and are not suitable for detecting semantic differences in all words in a corpus.%

In contrast, the proposed methods do not require any correspondence between corpora. Besides, they are efficient and effective. All they require are to compute the mean of contextualized word vectors and its norm for each word type. They do not require training nor have hyper-parameters to be searched for unlike previous methods. Nevertheless, they are effective even for corpus pairs whose sizes are skewed and for infrequent words. They are also capable of pinpointing word instances that have a meaning missing in one of the two corpora for comparison. For instance, in Sect.\,\ref{sec:eval}, they reveal that the word \textit{near} is one of the most typical words exhibiting a semantic difference between the native and non-native sub-corpora (approximately 10,000 and 100,000 words, respectively) of ICNALE~\cite{ishikawa} and that its most typical one out of the 11 \textit{near} occurrences in the native portion is ``\textit{it has \underline{near} impossible},'' which is interpreted as \textit{almost}; this usage does not appear at all in the 267 instances of \textit{near} in its counterpart.

The contributions of this paper are three-fold as follows: (i) We show for the first time that norms of the mean contextualized word vectors are good indicator for semantic differences; (ii) We give mathematical background to our rather intuitive methods; (iii) We actually reveal words that have semantic differences with their instances in native/non-native English and also 1800s/2000s English.

\section{Methods}\label{sec:methods}
We describe two methods, one for detecting words that have semantic differences in two corpora and one for extracting their typical instances. So far, we have often used the term \textit{word} abstractly to mean both \textit{word type} and \textit{word token}. Hereafter, for better understanding, we will distinguish between the two; we will use the term \textit{word type} to refer to word types and the term \textit{word instance} to a word token in context, which we assume is a whole sentence.%

\subsection{Detecting Semantic Differences}\label{subsec:method_for_detection}
To begin with, let us first note that the similarity between two words (tokens or types) are conventionally measured by the cosine similarity between the two word vectors (hereafter, for simplicity, word vectors will refer to contextualized ones unless otherwise noted). This is equivalent to measuring the word similarity based only on the directions of word vectors, or to assuming that all word vectors are normalized so that their Euclidean norms equal one. We follow this convention, hereafter.

\begin{figure}[t]%
\centering
    \includegraphics[scale=0.39]{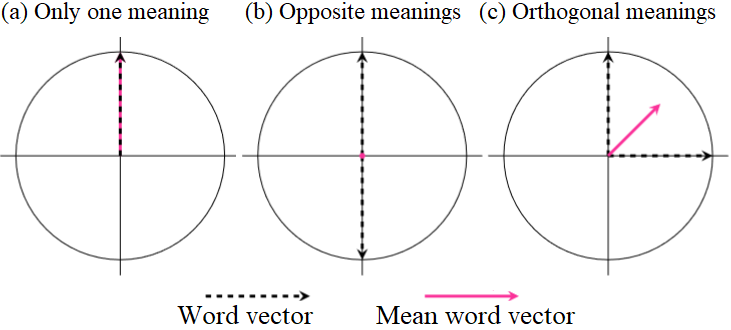}
\caption{Intuitive Illustration for Mean Norms.} \label{fig:von_mises_fisher}
\vspace{-0.7cm}
\end{figure}

Under this condition, any word vector appears on the unit hypersphere. As a special case of this, when the dimension of word vectors is two, word vectors appear on the unit circle as in the dashed arrows (vectors) in Fig.\,\ref{fig:von_mises_fisher}.%

With this preparation, we now examine the norm of the mean word vector for various cases. An extreme case would be that a word type is always used in the exact same context, and thus in the same meaning. Its word vectors appear on the same point on the unit hypersphere as in Fig.\,\ref{fig:von_mises_fisher}~(a). Then, its mean vector is always identical to the original word vectors, and thus its norm is also always one; recall all word vectors are normalized so that their norms equal one. The other extreme case would be that a word type is used in completely opposite meanings with the same frequency, which are represented by two opposite vectors as in Fig.\,\ref{fig:von_mises_fisher}~(b). In this case, its mean vector becomes the zero-vector with the zero norm. Other cases in between would give a norm between zero and one. For instance, two orthogonal vectors result in the mean word vector whose norm is $\frac{\sqrt{2}}{{2}}$ as in  Fig.\,\ref{fig:von_mises_fisher}~(c)\footnote{Addition of two orthogonal vectors produces a vector along the diagonal line with a norm of $\sqrt{2}$, and thus the norm of the mean word vector is $\frac{\sqrt{2}}{{2}}$.}.

The observations so far suggest that the wider meanings a word type cover in a given corpus, the shorter the norm of their mean word vector gets. This property of word vectors is the basis of the proposed methods.

To formalize the detection method, we will introduce the following symbols. We will denote a word vector by $\bm{x}$. Recall once again that $\|\bm{x}\|=1$ for all $\bm{x}$. We will also denote the mean vector of $\bm{x}$ and its norm by $\overline{\bm{x}}$ and $l$, respectively (i.e., $\overline{\bm{x}}\equiv \frac{1}{n}\sum_{i=1}^{n}\bm{x_i}$ where $n$ refers to the number of word instances of that word type in a given corpus). We will denote the two corpora for comparison by $S$ and $T$ (source and target\footnote{Source and target corpora would, for example, be native and non-native English corpora.}, respectively); for example, $l_S$ refers to the norm of the mean word vector of a word type obtained from the source corpus.

With these notations, the straight forward implementation of the above idea for measuring semantic differences would be taking the ratios $l_T/l_S$ for all word types appearing in two corpora; larger values of this indicate larger semantic differences (wider and narrower meanings in the source and target corpora, respectively).

In the proposed method, we use its extended version as our score function, which we call \textit{coverage}; we define coverage as
\begin{equation}
    c(S, T) = \frac{l_T(1-l_S^2)}{l_S(1-l_T^2)}, \label{eq:score}
\end{equation}
for which reason we will shortly describe in Subsect.\,\ref{subsec:method_background}. For the time-being, let us just notice that in the coverage, the norms $l_T$ and $l_S$ are respectively weighted by $1-l_S^2$ and $1-l_T^2$, which are based on its counterpart.

The procedure for detecting word types having semantic differences are as follows:\\%
\noindent
\textbf{Input}: source and target corpora $S$, $T$\\
\textbf{Output}: a list of words sorted in order of coverage\\
\vspace{-0.5cm}
\begin{enumerate}
\setlength{\parskip}{0cm}
\setlength{\itemsep}{0cm}
\item Vectorize all word instances in $S$ and $T$
\item For each word type, compute its mean vectors $\overline{\bm{x}}_S$ and $\overline{\bm{x}}_T$, and then its norms $l_S$ and $l_T$
\item Sort the word types by coverage defined by Eq.\,(\ref{eq:score}) in descending order
\item Output the sorted list
\end{enumerate}

\subsection{Extracting Typical Word Instances}\label{subsec:method_for_extraction}
\begin{figure}[t]%
\centering
    \includegraphics[scale=0.50]{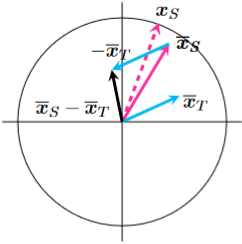}
\caption{Illustration for Difference of Mean Vectors.} \label{fig:mean_difference}
\end{figure}

We now turn our interest to extracting word instances having a meaning which is not, or seldom if ever, used in the target corpus. For this, we once again consider the illustrative unit circle shown in Fig.\,\ref{fig:mean_difference}. Fig.\,\ref{fig:mean_difference} shows two mean vectors of a word type obtained from the source and target corpora. Intuitively, word instances (or their word vectors) we are looking for now are those that are distant from the mean word vector for the target corpus (to make sure that their meanings are not or seldom used in it) and also that are near the mean word vector for the source corpus (to make sure that their meanings are indeed used in it). The dashed arrow $\bm{x}_S$ shown in Fig.\,\ref{fig:mean_difference} would be an example of this.

Fortunately, a difference of the two mean word vectors (i.e., $\overline{\bm{x}}_S - \overline{\bm{x}}_T$) will facilitate satisfying the conditions. Fig.\,\ref{fig:mean_difference} intuitively illustrates that the word vector $\bm{x}_S$ in the source corpus satisfies the two conditions. This corresponds to taking:%
\begin{equation}
    \cos(\overline{\bm{x}}_S - \overline{\bm{x}}_T, \bm{x}) = \frac{(\overline{\bm{x}}_s - \overline{\bm{x}}_T)^{\mathsf {T}}\bm{x}}{\| \overline{\bm{x}}_S - \overline{\bm{x}}_T\|\|\bm{x}\|} \label{eq:cos_diff_mean_x}
\end{equation}
Noting $\|\bm{x}\|=1$ and that $\|\overline{\bm{x}}_S - \overline{\bm{x}}_T\|$ is constant\footnote{Note that we are searching for $\bm{x}$ that gives a large value of Eq.(\,\ref{eq:cos_diff_mean_x}).} with respect to $\bm{x}$, Eq.\,(\ref{eq:cos_diff_mean_x}) reduces to
$ (\overline{\bm{x}}_s - \overline{\bm{x}}_t)^{\mathsf {T}}\bm{x}$. As in Subsect.\,\ref{subsec:method_for_detection}, we will adjust this cosine-based function by $1-l_S^2$ and $1-l_T^2$ to define another score function called \textit{representativeness} as%
\begin{equation}
    r(\bm{x}, S, T) =
    (\frac{1}{1-l_S^2}\overline{\bm{x}}_S - \frac{1}{l-l_T^2} \overline{\bm{x}}_T) ^{\mathsf {T}}\bm{x}, \label{eq:representativeness}
\end{equation}
to which we will give a mathematical background presently in Subsect.\,\ref{subsec:method_background}. %

%
%
%
%

The procedure for extracting word instances having a meaning which is not, or seldom used in the target corpus is as follows:\\%
\noindent
\textbf{Input}: source and target corpora $S$, $T$; a target word type $w$\\
\textbf{Output}: a list of word instances in $S$ sorted in order of representativeness\\
\vspace{-0.5cm}
\begin{enumerate}
\setlength{\parskip}{0cm}
\setlength{\itemsep}{0cm}
\item For $w$, compute the difference mean word vector ($\overline{\bm{x}}_S - \overline{\bm{x}}_T$)
\item For each word instance of $w$ in $T$, compute representativeness defined by Eq.(\,\ref{eq:representativeness})
\item Sort the word instances by representativeness in descending order
\item Output the sorted list
\end{enumerate}
The list obtained by swapping $S$ and $T$ is also helpful to investigate where the meaning difference comes from.

\subsection{Mathematical background}\label{subsec:method_background}
We now give a mathematical background to the proposed methods. Specifically, we show that the two score functions assume the von Mises-Fisher distribution behind word vectors (see the study~\cite{banerjee} for the detail of the distribution).

The von Mises-Fisher distribution is a probability density function for the random $d$-dimensional unit vector $\mathbf{x}$. It is defined as%
\begin{equation}
f(\mathbf{x};{\boldsymbol {\mu }}, \kappa) = z_\kappa \exp \left({\kappa {\boldsymbol {\mu }}^{\mathsf {T}}\mathbf {x} }\right). \label{eq:vfm}
\end{equation}
The parameters $\boldsymbol {\mu}$ ($\|\boldsymbol {\mu }\|=1$) and $\kappa$ ($ \kappa \geq 0$) are respectively the mean direction and concentration parameter. The constant $z_\kappa$ is the normalization constant depending on $\kappa$.
It can be regarded as akin to the isotropic Gaussian distribution of the hypersphere. It is commonly used to process directional data as in the present paper.

In our case, the unit vector $\mathbf{x}$ of the von Mises-Fisher distribution is the word vector $\bm{x}$. It follows that the word vector $\bm{x}$ distributes isotropically around the mean direction $\boldsymbol{{\mu}}$ with the concentration $\kappa$. Then, $\kappa$ is interpreted as the concentration of word meanings of the corresponding word types. In turn, the ratio $\kappa_T/\kappa_S$ measures the coverage of the meanings of that word type in the target corpus compared to those in the source corpus.

To examine the ratio $\kappa_T/\kappa_S$, one needs to estimate $\kappa$. \cite{banerjee} show a simple approximate solution of its maximum likelihood estimate is:%
\begin{equation}
    \kappa \approx \frac{l(d-l^2)}{1-l^2}, \label{eq:kappa}
\end{equation}
where $l$ is the norm of the mean vector as defined in Subsect.\,\ref{subsec:method_for_detection} while $d$ denotes the dimension of the unit vector $\mathbf{x}$. Then the ratio is approximated to%
\begin{equation}
    \frac{\kappa_T}{\kappa_S} \approx \frac{
                                            \frac{l_T(d-l_T^2)}{1-l_T^2}}{
                                    \frac{l_S(d-l_S^2)}{1-l_S^2}
                                }   \label{eq:kappa_ratio}
\end{equation}
Using\footnote{For example, $d=1024$ when 'bert-large-uncased' is used as a vectorizer while $l\in[0, 1]$.} $d \gg l$, it is further approximated to%
\begin{equation}
    \frac{\kappa_T}{\kappa_S} \approx \frac{l_T(1-l_S^2)}{l_S(1-l_T^2)}, 
\end{equation}
which is identical to our score function \textit{coverage}.

For the representativeness defined by Eq.\,(\ref{eq:representativeness}),  we can show that it is equivalent to examining the log likelihood ratio of the probability density function, which compares how probable the given $\bm{x}$ is in the two corpora. It is given by
\begin{eqnarray}
    \mathrm{LLR} &=& \log
                    \frac{
                        z_{\kappa_S} \exp \left({\kappa_S {\boldsymbol {\mu }_S}^{\mathsf {T}}\bm {x} }\right) 
                    }
                    {
                        z_{\kappa_T} \exp \left({\kappa_T {\boldsymbol {\mu }_T}^{\mathsf {T}}\bm {x} }\right)
                    }  \nonumber \\
                &=& \log \frac{z_{\kappa_T}}{z_{\kappa_S}} + (\kappa_T {\boldsymbol {\mu }_T} - \kappa_S {\boldsymbol {\mu }_T})^{\mathsf {T}} \bm{x}. \label{eq:llr}
\end{eqnarray}
The maximum likelihood estimate of $\boldsymbol{\mu}$ is given by $\boldsymbol{\mu} = \frac{\overline{\bm{x}}}{l}$~\cite{banerjee}. Here, note that the second term in the second line only matters with respect to $\bm{x}$. Then, putting this and Eq.\,(\ref{eq:kappa}) into the second term results in%
\begin{equation}
 (\frac{d-l_{S}^2}{1-l_{S}^2}{\overline{\bm{x}}_S} -  \frac{d-l_{T}^2}{1-l_{T}^2}{\overline{\bm{x}}_T} )^{\mathsf {T}}\bm{x}
\end{equation}
The approximations $d - l_{T}^2 \approx d - l_{S}^2$ for $d \gg l_{T}^2$ and $d \gg l_{S}^2$ give the score function \textit{representativeness}. Note the coarse approximation $\kappa \approx l$ would give the naive score functions originally introduced in Subsect.\,\ref{subsec:method_for_detection} and \ref{subsec:method_for_extraction}.

\section{Evaluation}\label{sec:eval}
\subsection{Data and Conditions}\label{subsec:data}
In this section, we detect word types having semantic differences and extract their word instances using the proposed methods to evaluate their effectiveness. Specifically, we compare the following two corpus pairs: native and non-native speaker English; 1800s and 2000s English. We use ICNALE~\cite{ishikawa} and the cleaned version~\cite{alatrash} of COHA~\cite{davies} for the former and latter, respectively. Table~\ref{tab:stats} shows their sizes.

\begin{table}[b]%
\begin{center}
\begin{tabular}{l|cc|cc}
\hline
Corpus & \# tokens \\
\hline
ICNALE Native & 97,899 \\
ICNALE Non-native &  986,764\\
COHA 1800s & 111,048,657  \\
COHA 2000s & 68,678,659 \\
\hline
\end{tabular}
\end{center}
\vspace{-0.2cm}
\caption{Sizes of Corpora for Evaluation.}\label{tab:stats}
\vspace{-0.3cm}
\end{table}%



ICNALE consists of essays written by native and non-native speakers of English. As a non-native sub-corpus, we use the essays labelled as either China, Indonesia, Japan, Korea, Taiwan, and Thailand. The essay topics are written on either (a) \textit{It is important for college students to have a part-time job.} or (b) \textit{Smoking should be completely banned at all the restaurants in the country.} This means that the essay topics are common to the native and non-native sub-corpora while their sizes are considerably different as shown in Table~\ref{tab:stats}.


COHA provides texts published in between 1820s and 2010s. Accordingly, we use the texts in the corresponding periods. In COHA, 5\% of ten consecutive tokens every 200 are replaced by `@' due to copy right regulations. We exclude sentences containing this special token from our analysis. They also contain a wide variety of fixed labels such as citation information as in \textit{Produced from page scans provided by Internet archive}. These inevitably make the norm of the mean word vector longer for the words in them. Also, they can be noise in that words would not appear in the corpora (e.g., \textit{Internet} in 1800s). Similarly, proper names often collocate with fixed contexts such as movie scripts (e.g., \textit{John: Yes, it is.}). We exclude these noisy word types and proper nouns from the sorted list of word types\footnote{We manually exclude such words by consulting their typical word instances from the lists shown in Table~\ref{tab:coha} in the following section.}.

The other conditions in this evaluation are as follows. In all corpora, we only target tokens whose occurrences are more than ten. We use `bert-large-uncased'~\cite{devlin-etal-2019-bert} in the Hugging Face implementation. We only target tokens that are not split into multiple sub-words and that consist only of alphabetic letters.

\subsection{Comparison between Native and Non-Native English Corpora}\label{subsec:native_vs_non-native}
Table~\ref{tab:icnale} shows the 12 most semantically different word types with their typical word instances where the source and target are the native and non-native sub-corpora in ICNALE. Note that ``$S$:'' and ``$T$:'' in the typical word instance column denote that the corresponding word instances are extracted from the source and target corpora, respectively.

Table~\ref{tab:icnale} reveals the following three major reasons why the word types have wider meanings in the native sub-corpus: influence from essay prompt, idiomatic phrases, and differences in construction and part-of-speech (POS). We describe their details in this order below.%

\begin{table*}[t]%
\begin{center}
\begin{tabular}{ccccl}
\hline
$\log c(S, T)$ & Word type & $f_{S}$ & $f_{T}$ & Typical word instance \\
\hline
0.61 & completely & 48 & 1662 &  $T$: ESSAY PROMPT \\
0.54 & near & 11 & 267 & $S$: $\cdots$ it has become \textit{near} impossible to $\cdots$  \\
0.50 & country & 48 & 1707 &  $T$: ESSAY PROMPT \\
0.46 & concerned & 11 & 113 &   $T$: $\cdots$ as far as I'm \textit{concerned} $\cdots$ \\
0.45 & third & 11 & 348 &   $T$: Third, $\cdots$    \\
0.39 & period & 13 & 115 &  $S$: Period! \\
0.37 & first & 87 & 1512 &  $T$: First, $\cdots$\\
0.36 & place & 67 & 1764 &  $S$: $\cdots$ fall into \textit{place}  $\cdots$ / $\cdots$ bans that they have in \textit{place} $\cdots$\\
0.38 & course & 46 & 489 &  $T$: Of \textit{course} $\cdots$\\
0.36 & taking & 34 & 461 &  $S$: $\cdots$ \textit{taking}  a part time job is $\cdots$ \\
0.34 & hold & 16 & 111 &  $S$ $\cdots$  \textit{hold} down a job  $\cdots$\\
0.35 & knowledge & 16 & 574 & $S$: $\cdots$ \small \textit{knowledge} that smoking and passive smoking kill people $\cdots$\\
\hline
\end{tabular}
\end{center}
\vspace{-0.2cm}
\caption{Semantic Differences Found in Native ($S$ource) and Non-native ($T$arget) Sub-corpora in ICNALE.}\label{tab:icnale}
\vspace{-0.5cm}
\end{table*}%

\textbf{Influence from essay prompt}: Simply, many of the non-native speakers use one of the essay prompts \textit{Smoking should be completely banned at all the restaurants in the country.} as it is. To be precise, the entire phrase appears 39 times and only once in the non-native and native sub-corpora, respectively. This naturally makes the contexts of \textit{completely} and \textit{country} rather fixed in the non-native sub-corpus, resulting in their long norms of their mean word vectors.

\textbf{Idiomatic phrases}: More interestingly, Table~\ref{tab:icnale} reveals word types used in an idiomatic phrase or a phrasal verb that seldom appear in the non-native sub-corpus, including \textit{fall into \underline{place}}, \textit{in \underline{place}} (as in \textit{effective}), and \textit{\underline{hold} down a job} (as in \textit{manage to keep the job}). In the non-native sub-corpus, the writers often use \textit{place} to refer to physical locations while the native speakers also use it metaphorically including the idiomatic phrases. For \textit{hold}, it appears more than 100 times in the non-native sub-corpus, but none collocates with \textit{down}, directly suggesting that most non-native speakers do not use or know the phrasal verb that native speakers use (four out of the 16 instances of \textit{hold} appear in the phrasal verb). Instead, they often use it as a transitive verb as in \textit{hold a job}, which also frequently appear in the native sub-corpus.

Idiomatic phrases play the opposite role, too. Specifically, the non-native speakers repeatedly \textit{concerned} and \textit{course} in the idiomatic phrases as shown in Table~\ref{tab:icnale}. Surprisingly, they use the idiom \textit{of course} 87\% of the time while the native speakers often use \textit{course} to mean \textit{a set of classes}, which decreases the relative frequency of the idiomatic phrase (63\%). Although strictly, they are not idiomatic phrases, the non-native speakers \textit{first} and \textit{third} in a fixed phrase. Surprisingly again, for instance, the first 886 word instances of \textit{first} (sorted by $c(S, T)$) are actually \textit{First, $\cdots$}. It should be emphasized that the nationalities of the non-native speakers range over six countries and nevertheless, fixed expressions like these are common to them. This is beyond the scope of this paper, but it would be interesting to reveal the reasons behind.

\textbf{Differences in construction and POS}: These are represented by \textit{near} and \textit{knowledge}. The former only appears 11 times in the native sub-corpus and one of them is used to mean \textit{nearly} or \textit{almost} as in the typical word instance in Table~\ref{tab:icnale}. Manual investigation reveals that this usage does not appear at all in the non-native portion. This is an example of the robustness for low frequency instances. Namely, the proposed methods can detect semantic differences found in low frequency instances. This is true for the other example \textit{knowledge}, which appears only 16 times in the native sub-corpus. Out of the 16, according to $c(S, T)$, the top two typical word instance of this is the one used in an appositive construction as shown in Table~\ref{tab:icnale}. This usage is seldom used in the non-native sub-corpus; as far as we checked manually, only two were the case out of the 574 instances\footnote{We first searched the non-native sub-corpus for the pattern \textit{knowledge that}, obtaining 25 instances. We then looked into them, finding that 22 cases were used in a relative clause and 1 in an erroneous construction.}. This agrees with the general knowledge that appositive constructions are difficult for learners of English.

\subsection{Comparison between 1800s and 2000s English Corpora}\label{subsec:19th_and_21st}
Table~\ref{tab:coha} shows the list of word types having semantic differences, which follows the same format as Fig.\,\ref{tab:icnale}. The first and second halves of Table~\ref{tab:coha} show those having wider meanings in the 1800s (source) than in the 2000s (target) corpus and the opposite, respectively.

\begin{table*}[t]%
\begin{center}
\begin{tabular}{ccccl}
\hline
\multicolumn{5}{l}{$S$(ource): 2000s English, $T$(arget): 1800s English } \\
$\log c(S, T)$ & Word type & $f_{S}$ & $f_{T}$ & Typical instance \\
\hline
\hline 
1.19 & whore & 57 & 483 &  $T$: $\cdots$  military lofts, \textit{whore} the birds are trained ,  $\cdots$ \\
1.06 & rebounds & 18 & 342 &  $S$: 11.8 \textit{rebounds} / $T$: his heart \textit{rebounds}. \\
1.01 & teen & 44 & 849 & $T$: $\cdots$  is fif // \textit{teen} hundred dollars  $\cdots$ \\
0.96 & hitter & 40 & 303 & $S$: a switch \textit{hitter} / $T$: $\cdots$  with \textit{hitter} feelings$\cdots$ \\
0.92 & recession & 32 & 539 & $T$: direct approach or \textit{recession} \\
0.90 & pregnant & 343 & 2290 & $T$: $\cdots$ in \textit{pregnant} illustrations of this great truth $\cdots$ \\
0.88 & tuna & 13 & 464 & $T$: $\cdots$ in which we \textit{tuna} ourselves with the peoples $\cdots$ \\
0.86 & coma & 30 & 345 & $T$: $\cdots$  must have \textit{coma} from god . $\cdots$ \\
0.85 & quantum & 43 & 901 & $T$: $\cdots$ the usual \textit{quantum} of abuse $\cdots$ \\
0.84 & diner & 14 & 635 & $T$: $\cdots$ \textit{diner} a dix / if \textit{diner} was an apple  $\cdots$ \\
\hline
\multicolumn{5}{l}{$S$(ource): 1800s, $T$(arget): 2000s} \\
\hline
0.93 & systemic & 210 & 24 & $T$: \textit{systemic} inflammatory responses \\
0.91 & dynamo & 77 & 39 & $S$: a dynamo of 5000 horse power / $T$: She was a \textit{dynamo}. \\
0.83 & conversions & 36 & 90 &  $\cdots$  $\cdots$ \\
0.83 & trigger & 1222 & 337 & $T$: to \textit{trigger} the immune system. \\
0.81 & strikers & 17 & 205 & $T$: \textit{Strikers} on three. \\
0.78 & grille & 100 & 11 & $S$: the big rusty \textit{grille} / $T$: bar and \textit{grille}\\
0.76 & rotating & 333 & 29 & $S$: the \textit{rotating} motion / $T$: $\cdots$ \textit{rotating} the pelvis $\cdots$ \\
0.73 & champs & 82 & 60 & $S$: \textit{Champs} Elysees / $T$: national \textit{champs} \\
0.73 & spectrum & 618 & 272 & $S$: the light of the \textit{spectrum} / $T$: a broad \textit{spectrum} of items \\
0.72 & norm & 446 & 15 & $S$: the only \textit{norm} of law / $T$: income above the \textit{norm} \\
\hline
\end{tabular}
\end{center}
\vspace{-0.2cm}
\caption{Word Types Having Semantic Differences in 1800s and 2000s English in COHA.}\label{tab:coha}
\vspace{-0.5cm}
\end{table*}%

The word types are classified into the following three categories: transcription errors, differences in POS, and potential semantic shift. As before, We describe their details in this order below.%

\textbf{Transcription errors}: They are flagged as a semantic difference because they are seemingly incorrectly transcribed, mostly in the 1800s corpus as in the typical word instance of \textit{whore}, which should be \textit{where}; other examples include \textit{teen} (incorrectly split as in \textit{fif teen}, \textit{coma} for \textit{come}, and \textit{tuna} for \textit{tune}?). Transcription errors increase the variation in the contexts of a word type and in turn shortens the norm of its mean word vector. It is crucial to remove transcription errors to conduct accurate analyses. This is especially true for historical corpora of which text is transcribed (semi)-automatically. The evaluation results suggest that the proposed methods may be used to detect transcription errors.

\textbf{Differences in POS}: \textit{trigger} and \textit{rotating} fall into this category. The former is used often as a noun and a verb in the 1800s and 2000s corpora, respectively while the latter as an adjective/present participle and a gerund in the respective corpora.

\textbf{Potential semantic shift}: All other word types in Table~\ref{tab:coha} exhibit a semantic difference. A typical example is \textit{rebounds}. In the 1800s corpus, it refers to something rebounding physically while in the 2000s corpus, it acquires a meaning referring to an action in basketball. According to Wikipedia\footnote{\url{https://en.wikipedia.org/wiki/Basketball}. Accessed on 5th, May, 2023.}, basketball was first played in 1891 and thus there had not been this usage in middle 1800s or earlier. A similar case is \textit{systemic} of which typical word instance is \textit{systemic inflammatory responses}. According to \cite{akshay}, the disease called \textit{Systemic inflammatory response syndrome} was first described by Dr.\ William R.\ Nelson in 1983, which attests the acquisition of a new meaning (or rather a new usage) in the 1900s.

It would be difficult to prove their semantic shifts from the available information, but they are at least all interpretable. Examples include \textit{pregnant} (\textit{filled with meaning} vs \textit{a woman having a baby}), \textit{quantum} (\textit{a unit} vs \textit{quantum} in physics), and \textit{diner} (\textit{metal bars} vs \textit{an eating place}), to name a few.

\section{Discussion}\label{sec:discussion}
The evaluation results in Sect.\,\ref{sec:eval} show that the concentration parameter $\kappa$, which is directly related to the norm of the mean word vector, is a good indicator for semantic differences. The differences shown in Table~\ref{tab:icnale} and Table~\ref{tab:coha} are all interpretable and some of them are indeed of meaning; after having seen few extracted typical word instances, we were able to tell, in most cases, where the difference(s) in the word type in question came from.%

\begin{figure}[t]%
\centering
\begin{center}
\begin{picture}(70, 128)
    \put(-75, 0){\includegraphics[scale=0.48]{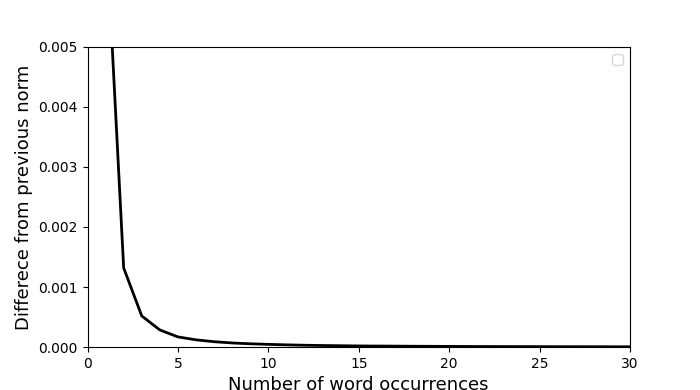}}
\end{picture}
\caption{Relationship between Number of Word Occurrences and Differences in Two
Consecutive Norms.}\label{fig:freq_vs_mean_norm}
\end{center}
\vspace{-0.5cm}
\end{figure}%

Here, it should be emphasized that semantic differences found in two corpora do not necessarily mean that the writers do not know/cannot use the missing meaning(s) (for non-native speakers) or that the word type has acquired/lost a meaning. It would require further investigations to confirm the argument. The proposed methods are rather suitable for obtaining new hypotheses about semantic differences in words or for supporting a hypothesis one already has.

Another advantage of the proposed methods is that they are computationally efficient. They require no training nor fine-tuning unlike the previous approaches as will be discussed in Sect.\,\ref{sec:related_work}. They solely rely on an off-the-shelf language model (BERT in our case), which is a large advantage in terms of implementation and development.

Correlated with this is that the proposed methods have almost no hyper-parameters except for the threshold for word frequency (word instances whose frequency is more than this threshold are the target of analysis). Fortunately, norms of mean word vectors are stable with respect to word frequency. To show this, we calculated norms of the mean word vector for each occurrence of each word type and then differences between the two consecutive values of the norms. Fig.\,\ref{fig:freq_vs_mean_norm} shows the results where the horizontal and vertical axes denote the number of occurrences of word types and the norm differences averaged over all word types. Fig.\,\ref{fig:freq_vs_mean_norm} shows that after around five occurrences, the average norm difference becomes almost zero, meaning that the norm of the mean vector is almost constant. Considering this, setting the frequency threshold to ten just as in the evaluation in Sect.\,\ref{sec:eval} is not a bad choice. This stability of the norm enables the propose methods to discover semantic differences in infrequent instances. It should be emphasized that only one word instance would be enough to proof that a meaning exists in a word type as in the \textit{near} example in Table~\ref{tab:icnale} (while the opposite does not hold).

It should be also emphasized that its robustness for the low frequency problem comes from the use of contextualized word vectors via a large language model. Even if the source and target corpora are small, the obtained word vectors should be statistically reliable considering the language model is trained on a large corpus. In contrast, the previous methods based on non-contextualized word vectors inevitably suffer from the low frequency problem because non-contextualized word vectors are learned from the input corpora.


As having discussed, the proposed methods are simple and efficient, but at the same time effective in discovering semantic differences found in words. All these nice properties come from the assumption of the von Mises-Fisher distribution behind word vectors. Although this assumption has its limitations theoretically, it works well practically as we have seen in Sect.\,\ref{sec:eval}.

\section{Related Work}\label{sec:related_work}
Linguists (e.g., \cite{fujimaru,mcenery}) often use frequency-based methods to discover differences in words in two corpora. Because they only consider superficial frequency counts, it requires more sophisticated methods to conduct deeper analyses into semantic differences.


The use of non-contextualized word vectors is the major approach to semantic difference detection. For diachronic analysis, \cite{kim-etal-2014-temporal} propose setting word vectors obtained from the previous time to initial word vectors of the next. For the same purpose, \cite{Kulkarni} and \cite{hamilton-etal-2016-diachronic} propose methods for discovering semantic differences by aligning words in two corpora. These alignment-based methods make a strong assumption that words are linearly aligned between two corpora, which does not necessarily hold in any corpus pair (e.g., comparison between native and non-native speakers). \cite{takamura-etal-2017-analyzing,kawasaki-etal-2022-revisiting} extends this approach to discovering semantic differences across languages while they require a word-alignment dictionary across languages.

\cite{yao} avoid the problem in word alignment by learning word vectors and alignment simultaneously. Their method has sensitive hyper-parameters that needs to be tuned, which results in a complex combinatorial optimization problem~\cite{aida-etal-2021-comprehensive}. \cite{dubossarsky-etal-2019-time} propose a method for detecting semantic differences by simultaneously optimizing multiple word vectors. While this method does not require linear transformations nor extensive hyper-parameter search, it requires a list of target words, which is not realistic in practical uses. \cite{aida-etal-2021-comprehensive} extends \cite{dubossarsky-etal-2019-time}'s method by optimizing multiple context vectors together with multiple word vectors. These non-alignment-based methods, however, still make the assumption that word vectors and/or context vectors are close to each other in two corpora. The task of detecting semantic differences is to find words that are not aligned well in terms of their meanings in two corpora, and thus methods requiring no assumption about the relation between word vectors in two corpora are preferable.

\cite{gonen-etal-2020-simple} propose a method that does not make such assumptions based on nearest neighbors obtained by non-contextualized word vectors, which makes it applicable to any pair of corpora. At the same time, it suffers from the bias in corpus sizes and the low frequency problem.

Some researchers try to use contextualized word vectors for semantic difference detection. \cite{hu-etal-2019-diachronic,giulianelli-etal-2020-analysing,kobayashi-etal-2021-analyzing} automatically group contextualized word vectors obtained to predict word meanings and then compare the results to detect semantic differences. Predicting word meanings is another difficult task itself.  Also, it is computationally costly to train a classifier or conduct clustering for every single word type found in corpora.


\section{Conclusions}\label{sec:conclusions}
In this paper, we have proposed using the norms of mean word vectors to detect semantic differences with their typical instances. The proposed methods do not require the assumptions concerning words and corpora for comparison that the previous methods do. The only assumption is that word vectors follow the von Mises-Fisher distribution. Accordingly, the proposed methods are applicable to corpus pairs such as native and non-native English corpora where the assumptions of the previous methods do not hold. Also, they are simple and efficient in that they do not require training nor extensive hyper-parameter search. With the methods, we have actually discovered semantic differences in native and non-native English corpora and also in historical corpora. We have revealed that they are effective even for infrequent word instances and also for corpora whose sizes are considerably different.


\bibliographystyle{acm}
\bibliography{anthology,custom}
\end{document}